\documentclass[10pt,twocolumn,letterpaper]{article}

\usepackage{cvpr}              

\usepackage{graphicx}
\usepackage{amsmath}
\usepackage{amssymb}
\usepackage{multirow}
\usepackage{setspace}
\usepackage{bm}
\usepackage[ruled]{algorithm}
\usepackage[dvipsnames,table,xcdraw]{xcolor}
\usepackage{pifont}
\usepackage{overpic}
\usepackage{enumitem}

\usepackage[pagebackref,breaklinks,colorlinks,linkcolor=red,citecolor=RoyalBlue]{hyperref}

\newcommand{\figref}[1]{Fig.~\ref{#1}}%
\newcommand{\secref}[1]{Sec.~\ref{#1}}
\renewcommand{\eqref}[1]{Eqn.~(\ref{#1})}

\newcolumntype{x}[1]{>{\centering\arraybackslash}p{#1pt}}

\newlength\savewidth
\newcommand{\tablestyle}[2]{\setlength{\tabcolsep}{#1}\renewcommand{\arraystretch}{#2}\centering\footnotesize}
\makeatletter\renewcommand\paragraph{\@startsection{paragraph}{4}{\z@}
  {.5em \@plus1ex \@minus.2ex}{-.5em}{\normalfont\normalsize\bfseries}}\makeatother

\newcommand{\myPara}[1]{\vspace{.05in}\noindent\textbf{#1}}

\def\cvprPaperID{xxxx} 
\def\confName{CVPR}
\def\confYear{2023}

\begin{document}

\title{Towards Spatial Equilibrium Object Detection}

\author{Zhaohui Zheng \qquad Yuming Chen \qquad
  Qibin Hou\thanks{Corresponding author} \qquad
  Xiang Li \qquad Ming-Ming Cheng \\
  TMCC, CS, Nankai University\\
}
\maketitle

\begin{abstract}
Semantic objects are unevenly distributed over images.
In this paper, we study the spatial disequilibrium problem of modern object detectors
and propose to quantify this ``spatial bias'' by measuring the detection performance over zones.
Our analysis surprisingly shows that the spatial imbalance of objects has a great impact on the detection performance, limiting the robustness of detection applications.
This motivates us to design a more generalized measurement, termed Spatial equilibrium Precision (SP),
to better characterize the detection performance of object detectors.
Furthermore, we also present a spatial equilibrium label assignment (SELA)
to alleviate the spatial disequilibrium problem by injecting the prior spatial weight into the optimization process of detectors.
Extensive experiments on PASCAL VOC, MS COCO, and 3 application datasets on face mask/fruit/helmet images demonstrate the advantages of our method.
Our findings challenge the conventional sense of object detectors and show the indispensability of spatial equilibrium.
We hope these discoveries would stimulate the community to rethink how an excellent object detector should be.
All the source code, evaluation protocols, and the tutorials are publicly available at \url{https://github.com/Zzh-tju/ZoneEval}.

\end{abstract}

\section{Introduction}\label{intro}

\begin{figure}[!t]
  \centering
  \includegraphics[width=0.46\textwidth]{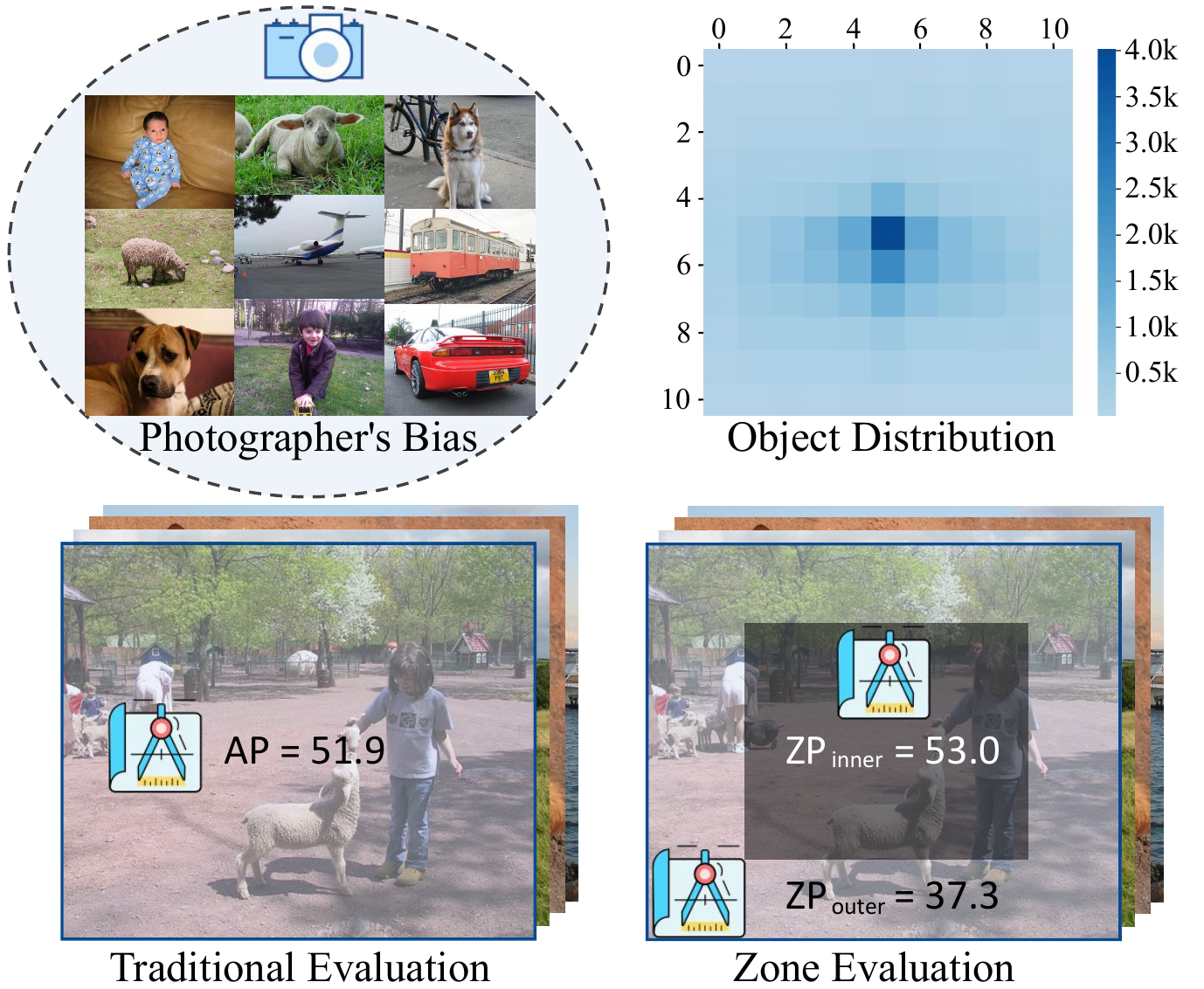}\\
  \caption{Top row: Photographer's bias causes spatial imbalance
    of object distribution.
    We count the center points of all the ground-truth boxes.
    The images are divided into $11\times11$ zones.
	Bottom row: The detection performance is measured by
    the traditional evaluation (Average Precision) and
    our zone evaluation (ZP, the average precision constrained in the zone).
    We show there is a large performance gap between zones.
    The results are reported by GFocal \cite{gfocal} on the VOC 2007 test set.
  }
  \label{fig:photographer-bias}
\end{figure}

Object detection, as one of the most popular vision tasks,
has been explored deeply over the past two decades
\cite{fasterrcnn,SSD,yolov3,DETR}.
Many endeavors have been made to push the benchmark ranking to a new level
\cite{cascadercnn,yolov4,yolov5,gfocal,VFNet,deformabledetr,dndetr}.
This paper does not aim to present a new object detector but
reveals the spatial disequilibrium problem in object detection,
which is mainly caused by the spatial imbalance of object distribution,
as shown at the top of \figref{fig:photographer-bias}.
This refers to the photographer's bias, which has been widely demonstrated to be intrinsic in the human's
biological visual system~\cite{tatler2007central,tseng2009quantifying}.
The photographers are more likely to place the important and
interesting objects at the center of the viewfinder.
It is natural, saliency-driven, related to the viewing strategy,
and found to be present in most of the existing visual datasets
\cite{torralba2011unbiased,oksuz2020imbalance,voc,coco,openimages,objects365}.
The majority of objects are intensively distributed over the middle zones.
This motivates us to investigate whether such photographer's bias affects
the detection performance and, if so, how much.

The core of this paper is a new zone evaluation protocol,
which attempts to measure the detector's performance over zones.
The new zone precision (ZP) calculates the common Average Precision within a designated zone,
where only the boxes whose centers lie in the zone are considered.
With the concept of ZP, it is possible to analyze the
spatial disequilibrium problem of various detectors over different image zones.
As shown in \figref{fig:photographer-bias},
the ZP gap is 15.7 between the inner zone and the outer one.
This performance gap indicates that the detector cannot perform uniformly
across the zones, which we refer to ``spatial bias".
It can be concluded from the experimental results that the spatial bias
is correlated to the spatial object distribution.
As a result, when the object distribution satisfies the centralized
photographer’s bias,
the detector will behave much better in the central zone
than in the outside zones.
This makes the traditional metrics, such as the popular Average Precision (AP),
actually be inflated,
heavily rely on the detection performance in a small central zone,
and hence fail to evaluate the spatial robustness of object detectors.

\begin{figure}[!t]
  \centering
  \small
  \begin{overpic}[width=0.46\textwidth]{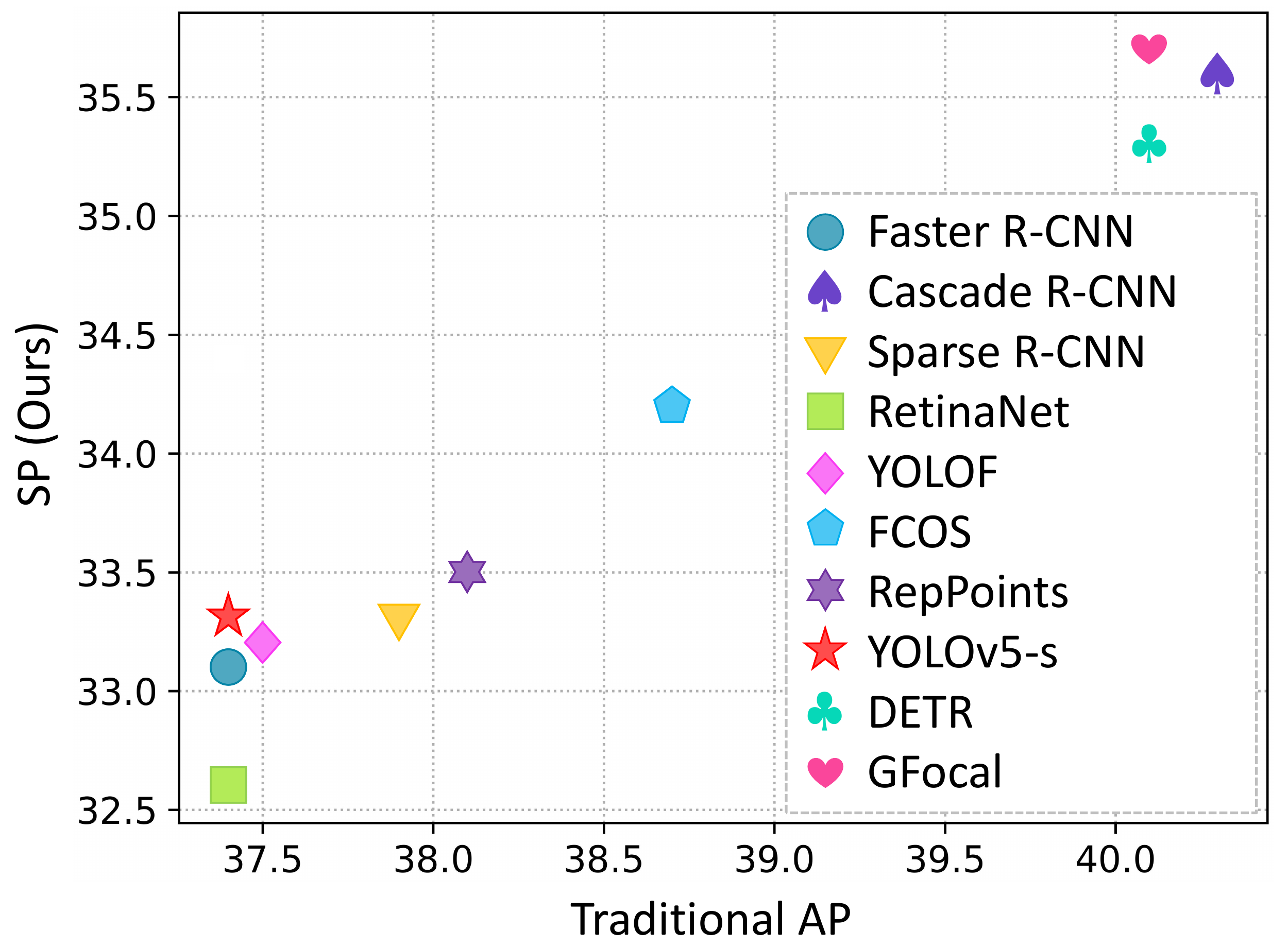}
	\put(88.5, 54.5){\cite{fasterrcnn}}
	\put(91.5, 50.0){\cite{cascadercnn}}
	\put(89.2, 45.3){\cite{sparsercnn}}
	\put(83.3, 40.4){\cite{lin2017focal}}
	\put(77.7, 35.9){\cite{yolof}}
	\put(76.0, 31.3){\cite{FCOS}}
	\put(83, 26.6){\cite{reppoints}}
	\put(82.5, 21.8){\cite{yolov5}}
	\put(76.1, 17.4){\cite{DETR}}
	\put(78.2, 12.8){\cite{gfocal}}
  \end{overpic}
  \caption{Spatial equilibrium Precision (SP) vs.
    traditional Average Precision (AP) for various object detection methods
    on COCO.
    Our SP considers the spatial equilibrium of detectors.
  }\label{fig:benchmark}
\end{figure}
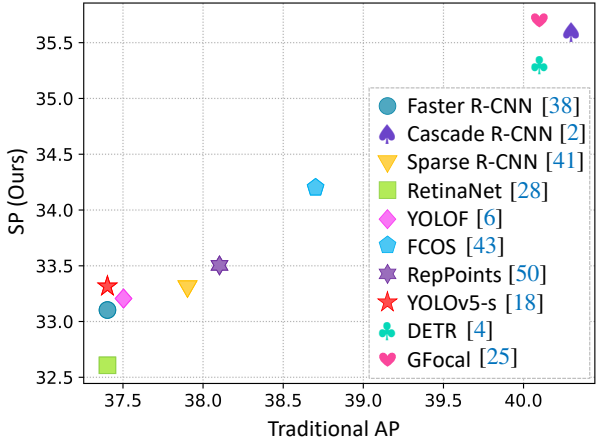

Driven by these findings, we present a new overall metric based on ZP,
called Spatial equilibrium Precision (SP),
to evaluate the object detectors in a spatial equilibrium manner.
More than the traditional AP,
SP can reflect the spatial robustness of detectors and
reduce the over-reliance on the small central zone.
With SP, we show surprising results where the benchmark ranking changes.
As shown in \figref{fig:benchmark},
YOLOv5-s \cite{yolov5} produces the lowest traditional AP among these detectors.
However, a higher SP score is achieved by YOLOv5-s since it performs
much better in terms of spatial equilibrium,
even on par with Sparse R-CNN~\cite{sparsercnn} which gets a traditional AP of $37.9\%$.
These results challenge our conventional understanding of the object detectors
and the spatial equilibrium should not be overlooked.

Moreover, as a preliminary attempt,
we propose a Spatial Equilibrium Label Assignment (SELA)
to re-balance the sampling process over the zones.
It successfully shrinks the performance gap between the central zone and
the ones near the image borders,
achieving a much better spatial equilibrium.
Experimental results on PASCAL VOC \cite{voc} and MS COCO \cite{coco}
support our analysis and conclusions.
We also conduct experiments on 3 application datasets,
including face mask/fruit/helmet images.
We hope our work has implications for promoting the comprehensiveness
and robustness of object detection.

\section{Related Work}\label{sec:related}

\subsection{Data Imbalance Problem}

Data imbalance is a common problem in object detection.
Let $\{X,G\}=\{x_i,g_i\}_{i=1}^{n}$ be a collection of sample-label pairs,
where each sample $x_i$ has a set of ground-truth labels $g_i$.
The model training is conducted on the subset of $\{X,G\}$,
and the network glances through the training set at each training epoch.
The data imbalance problems are usually related to the inherent properties
of $\{X,G\}$.
In the literature of object detection,
there are mainly two widely discussed imbalance problems.

\myPara{Class imbalance problem.}
In this case, the sample $X$ is composed of multiple subsets
$X_1,X_2,\cdots,X_c$ according to the class division,
where the number of samples is imbalanced across $c$ classes,
thereby yielding a long-tail distribution
\cite{zhang2021deep,ouyang2016factors,li2020overcoming,wang2021adaptive}.
The class imbalance problem naturally causes imbalanced sampling
during training,
hindering the classification performance for those tail classes.
Re-sampling strategies \cite{Kang2020Decoupling,mahajan2018exploring}
and cost-sensitive learning \cite{zhou2005training,cui2019class}
are the mainstream paradigms for class re-balancing.

\myPara{Foreground-background sampling imbalance.}
This imbalance is also derived from the data itself.
A large number of anchor points are tiled on the background area,
which are naturally sampled to be the negatives and hence dominant
the gradient flows.
In this case, $X$ can be split into $X_{n}$ and $X_{p}$,
s.t., $X=X_{n}\bigcup X_{p}$.
The negative samples $X_{n}$ can be seen as the complementary set of
the positives $X_{p}$,
whose ground-truth labels are ``background'' without bounding box annotations.
The solutions to this problem are similar, including re-sampling,
\eg, OHEM \cite{OHEM}, Guided Anchoring \cite{wang2019region}
and IoU-balanced sampling \cite{librarcnn}, and cost-sensitive learning,
\eg, Focal loss \cite{lin2017focal}, GHM loss \cite{GHM}, and PISA \cite{PISA}.

In this work, we consider the photographer's bias
\cite{tatler2007central,tseng2009quantifying}
where objects tend to be photographed in the central zone of an image.
In this case, the sample $X$ can be divided into multiple subsets
according to the spatial zones, just like the class division.
Generally speaking,
spatial imbalance shares the similar characteristics to class imbalance.
The difference is that the latter has a long-tail distribution across classes,
while the former considers the uneven distribution of objects over spatial zones.

\subsection{Label Assignment}
As a key component of object detectors,
label assignment determines the positive and negative samples for detectors to learn.
Early works~\cite{yolov3,SSD,fasterrcnn,lin2017focal} mostly adopt a
fixed label assignment strategy,
\eg, the max-IoU assigner, which is popular for years.
Later, FCOS \cite{FCOS} introduces the center sampling mechanism to
select the samples near the center point of the objects as the positives,
and a scale assignment is used to assign the appropriate FPN-level
for the objects with different scales.

Recently, dynamic label assignment has attracted more and more
research attentions.
In FSAF \cite{FSAF}, a dynamic FPN-level selection module is proposed.
FreeAnchor \cite{zhang2019freeanchor}, SAPD \cite{SAPD},
Auto-Assign \cite{zhu2020autoassign}, MuSu \cite{MuSu},
and DW \cite{DW} all adopt
the prediction-guided loss weighting method,
which can be regarded as a soft label assignment strategy.
By modeling the positive and negative anchors as a Gaussian Mixture Model,
PAA \cite{PAA} utilizes the EM algorithm \cite{EM} to conduct label assignment.
OTA/SimOTA~\cite{OTA,YOLOX} considers label assignment as an
Optimal Transport Problem.
ATSS \cite{ATSS} reveals an important fact that the performance gap
between dense anchor-based and anchor-free detectors lies in label assignment.
It takes advantage of the statistical characteristics of all the objects,
and calculates the data-related IoU to assign the labels.
Based on ATSS, TOOD~\cite{TOOD} and DDOD~\cite{DDOD2021} study the
task-aligned and task-disentangle label assignment, respectively.

Among all the aforementioned methods,
they mainly consider the instance-level information,
\eg, the prediction quality, the prior knowledge of objects,
while neglecting the object distribution in the scene.
In our work, we take this into account and propose a spatial equilibrium
label assignment to tackle the spatial disequilibrium problem.
Our intention is to train spatial equilibrium object detectors.

\section{Zone Evaluation for Object Detection}\label{sec3}

As mentioned in \secref{intro},
detector trained on dataset with imbalanced object distribution
has different performance in different image zones.
In this section, we extend the traditional object detection evaluation methods
by introducing a more comprehensive zone evaluation process.

Given a test image $I$ and a set of evaluation metrics $\mathcal{M}$,
the classic evaluation methods simultaneously calculate the metrics for
all the detections and the ground-truths over the whole image.
The elements in $\mathcal{M}$ can be the COCO-style
AP (Average Precision)~\cite{coco},
mAP across 10 IoU thresholds,
or AP for the small/medium/large objects,
which have been widely used in object detection.
These traditional evaluation metrics measure the detection performance
over the whole image zone but consider nothing
about the spatial robustness of object detectors.

\begin{figure}[!t]
  \centering
  \includegraphics[width=\linewidth]{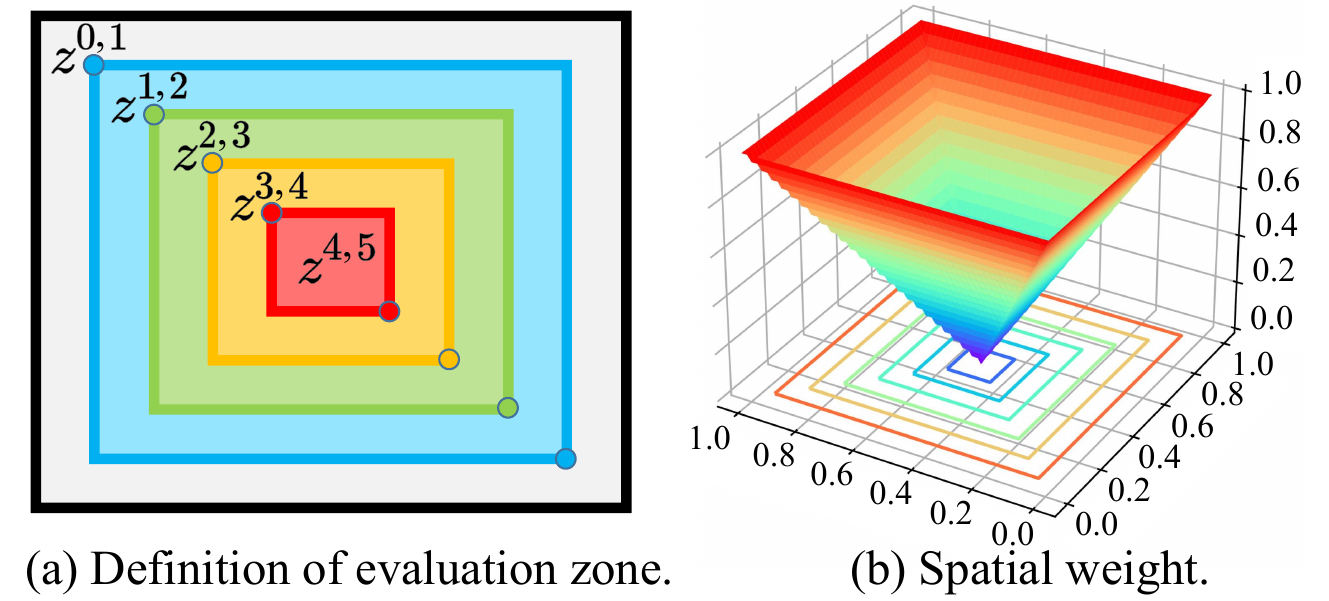}
  \caption{The definition of evaluation zones and the spatial weight
    in the normalized image space.
  }\label{fig:zone-weight}
\end{figure}

\myPara{Zone metric.}
Let $z^1,z^2,\cdots,z^n\in I$ be a set of disjoint image zones with union $I$.
We measure the detection performance for a specific zone $z^i$ by only considering the ground-truth objects and the detections whose centers lie in the zone $z^i$.
Then, for an arbitrary evaluation metric $m\in \mathcal{M}$, the evaluation process stays the same to the conventional ways, yielding $n$ zone metrics, each of which is denoted by $m^i$.

In practice, considering the widespread photographer's bias in the object detection datasets, as shown in \figref{fig:photographer-bias},
the evaluation zones are designed to be a series of annular areas:
\begin{equation}
    z^{i,j} = R_i\setminus R_j, \quad i<j
\end{equation}
where $R_i$ denotes a centralized region, which is given by:
\begin{equation}
    R_i = \mathrm{Rectangle}((r_iW,r_iH),((1-r_i)W,(1-r_i)H)),
\end{equation}
where $\mathrm{Rectangle}(p,q)$ represents the rectangle region with the top-left coordinate $p$ and the bottom-right coordinate $q$.
$W$ and $H$ denote the width and the height of the image and $r_i=\frac{i}{2n}, i\in\{0,1,\cdots,n\}$ controls the sizes of rectangles.
An illustration of the evaluation zones can be seen in \figref{fig:zone-weight}(a), where $n=5$.
We denote the range of the annular zone $z^{i,j}$ as $(r_i,r_j)$ for brevity.
And we denote the average precision (AP) in the zone $z^{i,j}$ as ZP$^{i,j}$.

\begin{figure}[!t]
	\centering
	\setlength{\tabcolsep}{1pt}
	\setlength{\abovecaptionskip}{3pt}
		\includegraphics[width=0.42\textwidth]{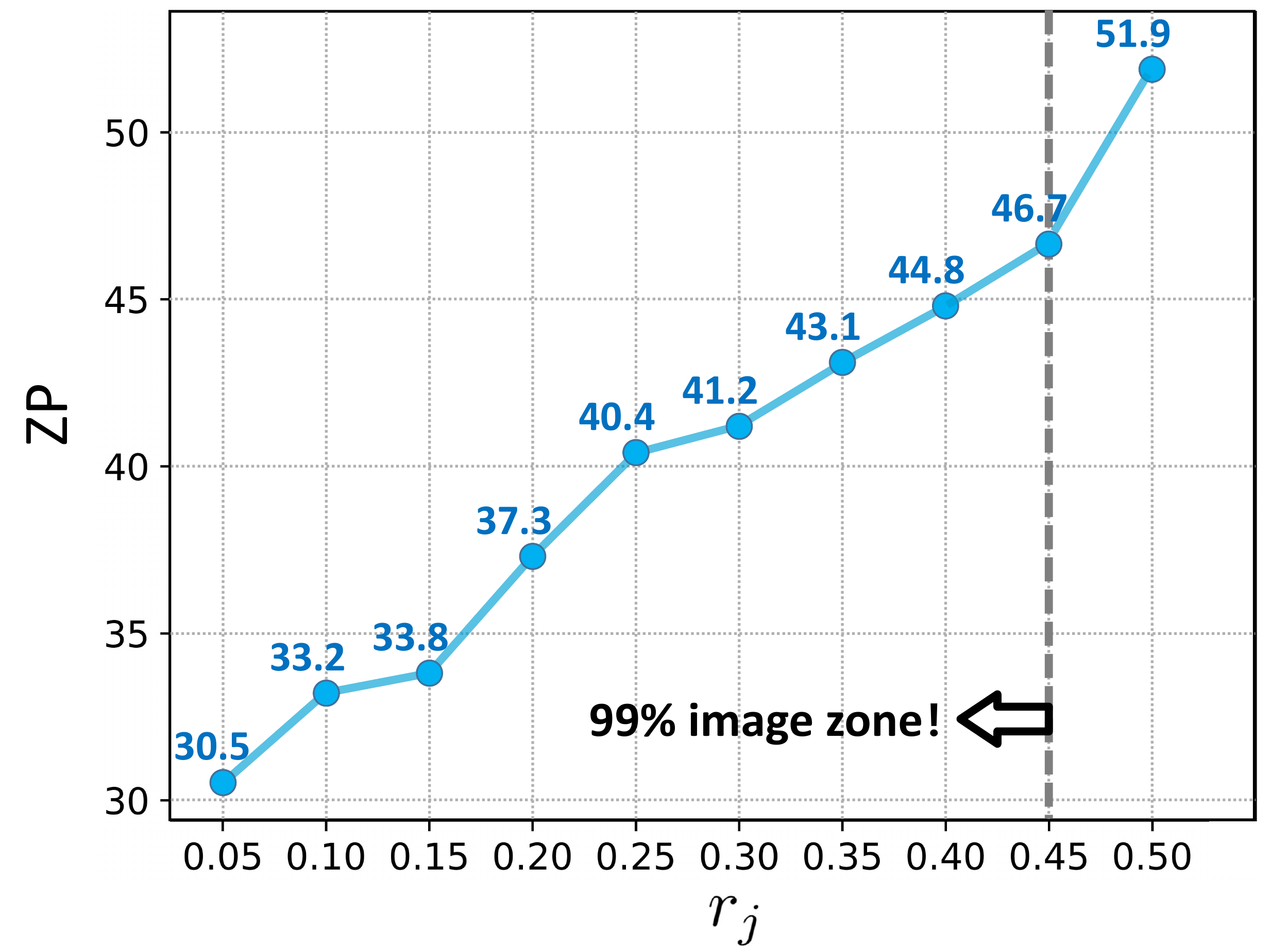}
	\caption{ZP scores against the zone range $(r_i,r_j)$, where $r_i$ is fixed as 0, and $r_j=0.05j$, $j\in\{1,2,\cdots,10\}$. We can see that even if a small central zone is excluded ($r_j=0.45$), which occupies only 1\% of the whole image area, the detection performance in the remaining 99\% zones can only produce 46.7 ZP, much lower than the traditional AP score over the whole image zone.
		}
	\label{fig:APcurve}
\end{figure}

\myPara{Rethinking traditional AP.}
Understanding the deficiency of the traditional AP is crucial for better applying object detectors to downstream applications.
It can be seen that the traditional AP is a special case in the proposed zone metric when $i=0$ and $j=n$, which corresponds to ZP$^{0,n}$.
This metric is difficult to capture the spatial bias, and is also dominated by the detection performance in a tiny zone.
To investigate its deficiency, we conduct evaluation on GFocal \cite{gfocal} with VOC 07+12 protocol.
We design a series of evaluation zones with range $(r_i,r_j)$, where the number of zones $n=10$, $r_i$ is fixed as $0$ and $r_j=0.05j$, $j\in\{1,2,\cdots,10\}$.
In a nutshell, all the evaluation zones can be regarded as hollowing the central zones $R_j$ in the full-map zone.
\figref{fig:APcurve} shows us quite surprising results that when $r_j<0.5$, all the ZPs are less than the traditional AP ($r_j=0.5$).
Notice that even if a small central zone is excluded ($r_j=0.45$), which occupies only 1\% of the whole image area, the detection performance in the remaining 99\% zones can only produce 46.7 ZP, much lower than the traditional one.
This exposes the defects of traditional metrics.
They are actually inflated and largely dominated by the detection performance in a small central zone.
Due to this fact, the traditional metrics can hardly reflect the real overall performance of the detectors.

\myPara{Spatial equilibrium metrics.}
Motivated by the aforementioned analysis, here, we attempt to propose a spatial equilibrium metric, which considers both the zone metrics and the zone areas to better reflect the detection performance.
Considering the normalized image space (square image with unit area 1), the spatial equilibrium metric for a given metric $m$ is given by:
\begin{equation}\label{eq:SP}
    \mathrm{S}m=\sum\limits_{i}\mathrm{Area}(z^i) \cdot m^i,
\end{equation}
where $\mathrm{Area}(\cdot)$ calculates the area of the zone.
In this way, the spatial equilibrium metric $\mathrm{S}m$ is a weighted sum of the zone metrics.
One may note that the spatial equilibrium metric is based on an assumption similar to the traditional metric, i.e., the detector performs uniformly in the zone.
The difference is, our spatial equilibrium metric applies this assumption to a series of smaller zones, rather than the full map for traditional metrics.

\begin{table}[htp!]
  \small
  \centering
  \setlength{\tabcolsep}{14pt}
  \caption{Notations of different evaluation metrics.}
  \vspace{-5pt}
  \begin{tabular}{ll}
    Notation & Descriptions \\ \hline
    ZP$^{i,j}$ & Average Precision (AP) in $z^{i,j}$. \\
    mZP$^{i,j}$ & mAP under a given IoU threshold in $z^{i,j}$. \\
    ZP$_{75}^{i,j}$ & AP$_{75}$ in $z^{i,j}$. \\
    SP & Spatial equilibrium metric for ZP. \\
    SP$_{75}$ & Spatial equilibrium metric for ZP$_{75}$. \\
  \end{tabular}
  \label{tab:notations}
\end{table}

\myPara{Notations.} For simplicity, the notations can be seen in Table \ref{tab:notations}.
In the experiments, SP should be considered as the most important metric for characterizing the detection performance.

\myPara{Measuring the discrete amplitude for zone metrics.} As the detection performance varies across the zones, we further introduce an additional metric to gauge the discrete amplitude among the zone metrics.
Given all the zone metrics for $m$, we calculate the variance of the zone metrics,
\begin{equation}
\sigma(m) = \frac{\sum_i(m^i-\bar{m})^2}{n},
\end{equation}
where $\bar{m}=\sum_im^i/n$ denotes the mean value of the zone metrics, and
$n$ denotes the number of zones.
Ideally, if $\sigma(m)=0$, the object detector reaches perfectly spatial equilibrium under the current zone division.
In this situation, an object can be well detected without being influenced by its spatial position.

\section{Spatial Equilibrium Label Assignment}\label{sec:SELA}

Re-sampling is one of the promising solutions to enhance the robustness of object detectors.
It is also widely used for alleviating the long-tail class imbalance.
For example, increasing sampling frequency for the tail classes can provide more training samples for the network to generalize better.
We, in this work, propose a simple yet effective Spatial Equilibrium Label Assignment (SELA) to relieve the issue of spatial disequilibrium.
We first introduce the key component of our method---the spatial weight.
We map the anchor point coordinate $(x^{a},y^{a})$ to a spatial weight $\alpha(x^{a},y^{a})$ by a spatial weighting function,
\begin{equation} \label{eq:spatial_weight}
    \alpha(x,y) = 2\max\left\{||x-\frac{W}{2}||_1\frac{1}{W}, ||y-\frac{H}{2}||_1\frac{1}{H}\right\},
\end{equation}
where $W$ and $H$ are the width and the height of the image.
The spatial weight has the following properties:
\begin{itemize}
    \item Non-negativity;
    \item Bounded by $[0,1]$;
    \item $\lim\limits_{(x,y)\rightarrow (\frac{W}{2},\frac{H}{2})}\alpha(x,y)=0$;
    \item When $p$ is located at the image border, we have $\lim\limits_{(x,y)\rightarrow p}\alpha(x,y)=1$.
\end{itemize}
An illustration of the spatial weight can be seen in \figref{fig:zone-weight}(b).
Next, we show how to use the spatial weight defined in Eq.~\ref{eq:spatial_weight} to improve the classic detectors.

\myPara{Usage of spatial weight.}
The spatial weight can be easily plugged into the existing label assignment algorithms.
The principle is simple and multi-optional.
The key idea is to consider the spatial weight as an additional constraint term when making the criterion rule for label assignment.
Since most of the label assignment algorithms have their own sophisticated implementations, in the following, we provide a specific application description of the classic ATSS \cite{ATSS}, simply because of its brevity.
Given the positive IoU threshold $t$, which is calculated by considering the statistical characteristics of the objects.
The ATSS criterion follows the same rule as the max-IoU assignment \cite{lin2017focal,yolov3,fasterrcnn},
\begin{equation}
    \textrm{IoU}(\bm{B}^{a},\bm{B}^{gt})\geqslant t,
\end{equation}
where $\bm{B}^{a}$ and $\bm{B}^{gt}$ denote the preset anchor boxes and the ground-truth boxes.
Our SELA process is represented as:
\begin{equation}\label{eq:SELA}
    \textrm{IoU}(\bm{B}^a,\bm{B}^{gt})\geqslant t-\gamma\alpha(x^{a},y^{a}),
\end{equation}
where $\gamma\geqslant0$ is a hyperparameter.
It can be seen that SELA relaxes the positive sample selection conditions for objects near the image borders.
Therefore, more anchor points will be selected as the positive samples for them.

\myPara{Discussions.}
Notice that the above simple application is actually a frequency-based approach, just like many of the class re-balance sampling strategies proposed for the long-tail class imbalance problem \cite{Kang2020Decoupling,mahajan2018exploring}.
This is not the unique option to pursue spatial equilibrium.
Cost-sensitive learning \cite{zhou2005training,cui2019class} is also worth considering.
We exploit the spatial weight to enlarge the loss weight for positive samples.
Thus, a larger gradient flow will be generated for the outer zones to make the network focus more on the objects away from the center.
Besides, there are some potentially promising solutions toward spatial equilibrium that deserve further study.
For example, designing an appropriate data augmentation, more specifically, increasing data augmentation to make up the insufficient sampling frequency for the objects near the image borders, might be a promising solution toward spatial equilibrium.
Also, we can actually design a skew spatial weight if the data has a skew object distribution.
We leave them for future research.

\begin{table}[htp!]
  \small
  \centering
  \setlength{\tabcolsep}{7pt}
  \caption{Brief information of the datasets for training and test.}
  \vspace{-7pt}
  \begin{tabular}{lcccc}
    Dataset    & Training & Test  & \#Classes & Source\\ \hline
    PASCAL VOC & 16,551   & 4,952 & 20 & \cite{voc}\\
    MS COCO    & 118K     & 5,000 & 80 & \cite{coco}\\
    Face mask  & 5,865    & 1,035 & 2  &  Kaggle\\
    Fruit      & 3,836    & 639   & 11 &  Kaggle \\
    Helmet     & 15,887   & 6,902 & 2  &  Kaggle \\
  \end{tabular}
  \label{tab:info}
\end{table}

\begin{figure}[!t]
	\centering
	\setlength{\tabcolsep}{1pt}
	\setlength{\abovecaptionskip}{3pt}
		\begin{overpic}[width=0.42\textwidth]{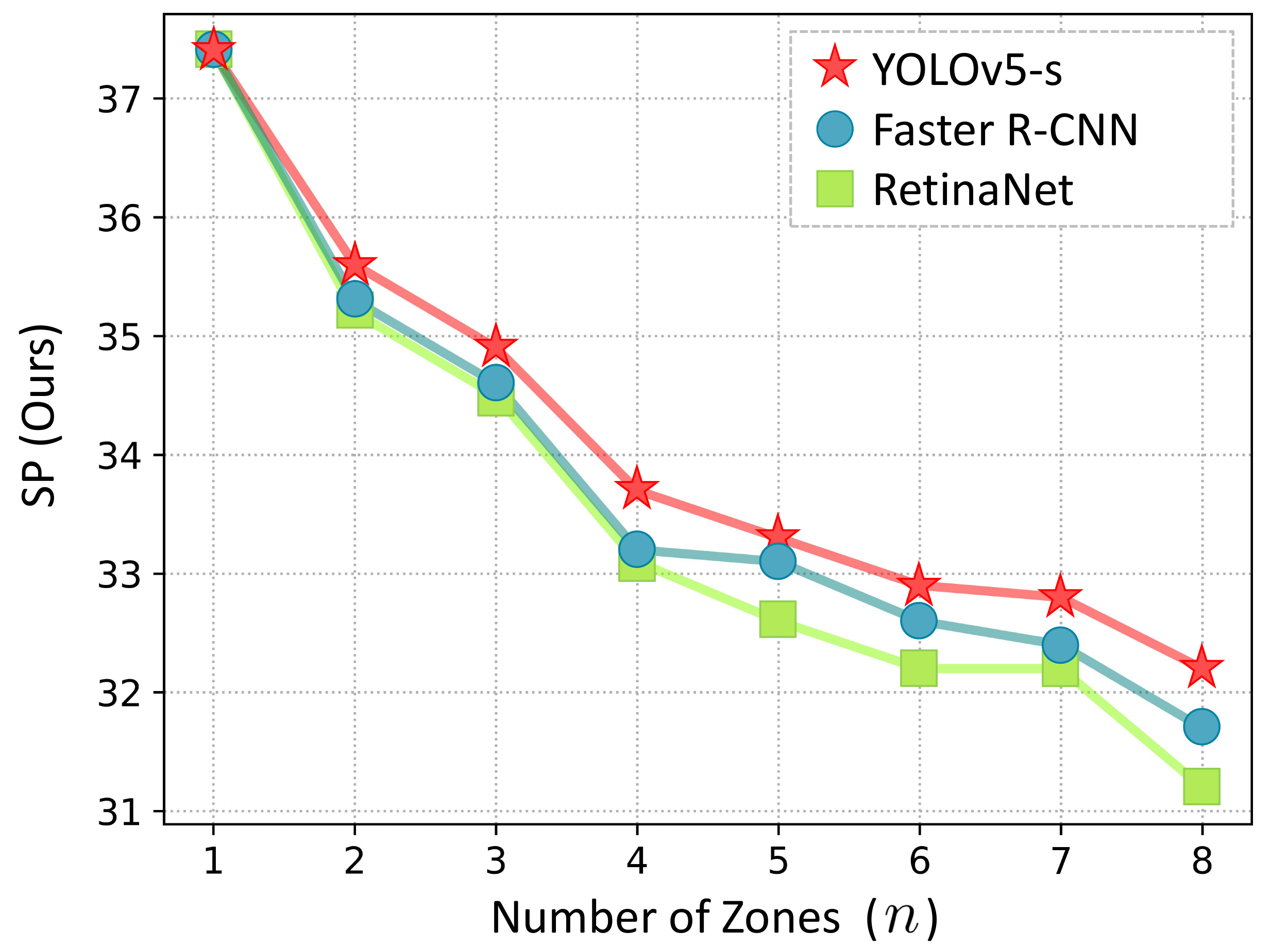}
		\scriptsize
		\put(82.8, 67.5){\cite{yolov5}}
		\put(88.5, 62.8){\cite{fasterrcnn}}
		\put(83.5, 58){\cite{lin2017focal}}
		\end{overpic}
	\caption{Spatial equilibrium Precision (SP) against the number of zones. $n=1$ denotes the traditional full map evaluation. We set $n=5$ by default. The results are reported on COCO 2017 val.
		}
	\label{fig:numberofzones}
\end{figure}
\section{Experiment} \label{sec:results}

We conduct extensive experiments on various object detectors and 5 detection datasets, including PASCAL VOC \cite{voc} MS COCO \cite{coco}, and 3 public application datasets (face mask/fruit/helmet images).
The experiments contain two parts, the zone evaluation and the SELA evaluation.

\subsection{Implementation Details}
For zone evaluation, we follow the standard Average Precision evaluation protocols.
All the object detectors we evaluate can be downloaded from MMDetection \cite{mmdetection} or their official websites.
To comprehensively evaluate the detectors and our SELA, various metrics are reported, including the SP, 5 zone metrics, the variance of 5 zone metrics, and the traditional metrics.
Unless otherwise stated, we adopt $n=5$ for the zone division, which can be seen in \figref{fig:zone-weight}(a).
The types of the evaluation metrics follow the standard protocols, i.e., AP (Average Precision) and AP$_{75}$, thereby yielding SP, SP$_{75}$, ZP, ZP$_{75}$, etc.

For the usage of datasets, we adopt the classical VOC 07+12 training and test protocols for PASCAL VOC.
The training set contains the union of VOC 2007 trainval and VOC 2012 trainval and the test set contains VOC 2007 test.
For MS COCO, we adopt COCO 2017 train (118K images) for training and COCO 2017 val (5K images) for evaluation.
Moreover, we adopt the VOC-style training and test protocols for 3 application datasets.
All the datasets we used are publicly available, and can be downloaded from their official websites or Kaggle.
The brief information is listed in Table \ref{tab:info}.
More details can be seen in the Appendix.

For SELA evaluation, the implementation is based on the popular dense object detector GFocal \cite{gfocal} with ResNet \cite{ResNet} backbone and FPN \cite{FPN} neck under the MMDetection \cite{mmdetection} framework.
We use ResNet-18 for VOC 07+12 and 3 application datasets, and adopt ResNet-50 for MS COCO.
The learning rate setting follows the linear scaling rule \cite{goyal2017accurate} according to the number of GPUs.
The training epochs are set to 12 for all the experiments.
All the other hyperparameters except $\gamma$ in Eq. \ref{eq:SELA} remain unchanged for a fair comparison.

\subsection{Analysis on Zone Evaluation}

\begin{table*}[!t]
  \centering
  \small
  \setlength{\tabcolsep}{8.5pt}
  \caption{Performance comparison among the existing popular object detectors.
    The spatial equilibrium precisions (SP), 5 zone precisions (ZP),
    the variance of ZPs, and the traditional metrics are reported.
    The results are reported on COCO 2017 val.
    \textbf{R}: ResNet \cite{ResNet}.
    \textbf{X}: ResNeXt-32x4d \cite{xie2017aggregated}.
    \textbf{PVT-s}: Pyramid vision transformer-small \cite{PVT}.
  }\vspace{-5pt}
  \begin{tabular}{lcccccccc}
	Detector & SP  & ZP$^{0,5}$  & Variance & ZP$^{0,1}$ & ZP$^{1,2}$ & ZP$^{2,3}$ & ZP$^{3,4}$ & ZP$^{4,5}$ \\ \hline
    DETR (\textbf{R}-50) \cite{DETR}
    & 35.3 & 40.1 & 26.9 & 29.8 & 36.2 & 39.8 & 39.1 & 45.7 \\
    RetinaNet (\textbf{PVT-s}) \cite{PVT}
    & 35.5 & 40.4 & 19.7 & 30.8 & 36.9 & 39.0 & 37.4 & 44.6 \\
    Cascade R-CNN (\textbf{R}-50) \cite{cascadercnn}
    & 35.6 & 40.3 & 18.7 & 30.9 & 36.6 & 39.2 & 38.6 & 44.2 \\
    GFocal (\textbf{R}-50) \cite{gfocal}
    & 35.7 & 40.1 & 14.4 & 30.9 & 37.2 & 39.1 & 38.3 & 42.5 \\\hline
    Cascade Mask R-CNN (\textbf{R}-101) \cite{cascadercnn}
    & 40.3 & 45.4 & 22.4 & 34.7 & 41.6 & 44.3 & 44.4 & 49.1 \\
    Sparse R-CNN (\textbf{R}-50) \cite{sparsercnn}
    & 40.6 & 45.0 & 21.6 & 35.8 & 41.9 & 43.4 & 44.0 & 50.3 \\
    YOLOv5-m \cite{yolov5}
    & 40.8 & 45.2 & 12.9 & 36.0 & 42.3 & 44.5 & 43.2 & 46.7 \\\hline
    Mask R-CNN (\textbf{Swin-T}) \cite{swin}
    & 40.9 & 46.0 & 15.4 & 36.8 & 41.7 & 44.1 & 43.5 & 49.0 \\
    Mask R-CNN (\textbf{ConvNeXt-T}) \cite{convnet}
    & 41.1 & 46.2 & 17.6 & 36.7 & 41.9 & 44.5 & 43.6 & 49.7 \\
    Cascade Mask R-CNN (\textbf{X}-101) \cite{cascadercnn}
    & 41.2 & 46.1 & 21.1 & 36.1 & 42.0 & 44.8 & 45.9 & 49.9 \\
    VFNet (\textbf{R}-101) \cite{VFNet}
    & 41.5 & 46.2 & 15.6 & 36.7 & 43.0 & 45.0 & 44.5 & 48.8 \\
    Deformable DETR (\textbf{R}-50) \cite{deformabledetr}
    & 41.6 & 46.1 & 23.2 & 36.3 & 42.6 & 45.6 & 45.1 & 51.2 \\
    Sparse R-CNN (\textbf{R}-101) \cite{sparsercnn}
    & 41.7 & 46.2 & 21.1 & 36.9 & 42.9 & 44.9 & 44.7 & 51.3 \\
    GFocal (\textbf{X}-101) \cite{gfocal}
    & 41.8 & 46.1 & 15.7 & 37.0 & 43.5 & 45.0 & 44.4 & 49.3 \\
  \end{tabular}
  \label{tab:detectors}
\end{table*}

\myPara{Number of Zones.}
We first investigate how the number of zones $n$
affects detectors' performance.
Recall that the zones are defined as a series of disjoint annular regions, and their union is the whole image.
A typical illustration ($n=5$) can be seen in \figref{fig:zone-weight}(a).
As shown in \figref{fig:numberofzones}, we plot the SP curves against the number of zones for YOLOv5-s \cite{yolov5}, Faster R-CNN \cite{fasterrcnn}, and RetinaNet \cite{lin2017focal}.
One can see that when $n=1$, our SP is identical to the traditional AP as the term $\mathrm{Area}(z^{0,n})=1$ in Eq. \ref{eq:SP}, which means that the detectors are assumed to perform uniformly in the whole image zone.
As $n$ increases, the requirements for spatial equilibrium become stricter and stricter.
When $n>1$, the performance gaps of these 3 detectors change.
And when $n\geqslant5$, the performance gap is distinguishable enough and the detector ranking becomes stable.
Therefore, considering the metric computations, we set $n=5$ by default and it should be noted that a large $n$ is also acceptable if a more rigorous spatial equilibrium is required.

\begin{table*}[!t]
  \centering
  \small
  \setlength{\tabcolsep}{4.2pt}
  \caption{Evaluation of SELA for the spatial equilibrium precisions (SP),
    5 zone precisions (ZP), the variance of ZPs,
    and the traditional metrics on 4 datasets,
    including VOC 07+12 and Face mask/Fruit/Helmet detection.
  }
  \begin{tabular}{l|c|cc|cc|cc|cc|cc|cc|cc|cc}

	\multirow{2}{*}{Dataset} & \multirow{2}{*}{SELA} &
    \multicolumn{2}{c|}{Overall} & \multicolumn{2}{c|}{$z^{0,5}$} &
    \multicolumn{2}{c|}{Variance} & \multicolumn{2}{c|}{$z^{0,1}$} &
    \multicolumn{2}{c|}{$z^{1,2}$} & \multicolumn{2}{c|}{$z^{2,3}$} &
    \multicolumn{2}{c|}{$z^{3,4}$} & \multicolumn{2}{c}{$z^{4,5}$}\\ \cline{3-18}
	&& SP & SP$_{75}$  & ZP & ZP$_{75}$ & ZP & ZP$_{75}$ & ZP & ZP$_{75}$ &
    ZP & ZP$_{75}$ & ZP & ZP$_{75}$ & ZP & ZP$_{75}$ & ZP & ZP$_{75}$ \\ \hline
    \multirow{2}{*}{VOC 07+12} &  & 37.2 & 40.3 &    51.9 & 56.0 &
                                    49.4 & 66.4 &    31.5 & 33.5 &
                                    37.7 & 41.3 &    40.1 & 43.6 &
                                    43.4 & 46.5 &    52.8 & 58.5 \\
                    &  \checkmark & 38.6 & 41.1 &    52.5 & 57.0 &
                                    37.9 & 58.8 &    33.9 & 34.3 &
                                    38.6 & 42.0 &    41.5 & 45.0 &
                                    43.3 & 47.7 &    52.5 & 57.8 \\ \hline
    \multirow{2}{*}{Face Mask} &  & 61.5 & 74.5 &    69.0 & 86.9 &
                                    17.9 & 31.5 &    55.3 & 65.4 &
                                    63.7 & 80.7 &    65.9 & 80.8 &
                                    65.9 & 76.0 &    66.8 & 76.2 \\
                    &  \checkmark & 62.8 & 76.5 &    69.1 & 87.5 &
                                    9.1  & 19.4 &    57.9 & 69.6 &
                                    65.5 & 81.8 &    65.9 & 81.8 &
                                    65.2 & 75.8 &    65.0 & 75.9 \\ \hline
    \multirow{2}{*}{Fruit}   &  &   67.0 & 76.1 &    74.7 & 84.1 &
                                    50.9 & 47.7 &    61.9 & 70.9 &
                                    67.9 & 77.8 &    68.3 & 76.5 &
                                    72.9 & 81.5 &    83.3 & 91.7 \\
                    &  \checkmark & 67.8 & 78.0 &    75.2 & 85.0 &
                                    37.9 & 28.6 &    64.1 & 76.5 &
                                    67.9 & 77.6 &    68.8 & 77.0 &
                                    72.4 & 81.1 &    82.2 & 90.8 \\ \hline
    \multirow{2}{*}{Helmet}   &  &  52.0 & 54.4 &    55.0 & 55.7 &
                                    3.2  & 9.0  &    50.1 & 52.4 &
                                    51.4 & 52.6 &    54.5 & 56.7 &
                                    54.7 & 59.9 &    52.1 & 58.2 \\
                    &  \checkmark & 52.8 & 55.9 &    55.5 & 56.7 &
                                    2.5  & 6.7  &    51.5 & 54.8 &
                                    52.0 & 53.5 &    54.9 & 58.0 &
                                    55.4 & 60.3 &    52.6 & 59.1 \\
  \end{tabular}
  \label{tab:datasets}
\end{table*}

\myPara{Evaluation of various object detectors.}
In Table \ref{tab:detectors}, we report the performance comparison for various popular detectors with the same level traditional metrics.
These detectors have different detection pipelines, e.g., one-stage dense detectors (RetinaNet \cite{lin2017focal}, GFocal \cite{gfocal}, VFNet \cite{VFNet}, YOLOv5 \cite{yolov5}), multi-stage dense-to-sparse detectors (R-CNN series \cite{fasterrcnn,cascadercnn,maskrcnn}), and sparse detectors (DETR series \cite{DETR,deformabledetr} and Sparse R-CNN \cite{sparsercnn}).
One can see that GFocal and DETR achieve the same traditional AP 40.1, i.e., ZP$^{0,5}$.
However, GFocal performs better in the outer zones $z^{0,1}$ and $z^{1,2}$, which occupy 64\% image areas.
Our spatial equilibrium metrics take the spatial equilibrium into account and show a 0.4 SP superiority of GFocal over DETR.
Similarly, Mask R-CNN (\textbf{Swin-T}) has a comparable traditional AP to GFocal (\textbf{X}-101).
One can see that GFocal performs sightly better than Mask R-CNN in the zone $z^{0,1}$ and $z^{4,5}$,
while significantly better in the 3 zones $z^{1,2}$, $z^{2,3}$, and $z^{3,4}$, and therefore shows a $0.9$ SP gap to Mask R-CNN.
One more interesting is the sparse detectors, e.g., DETR series and Sparse R-CNN tend to produce a large variance, while the one-stage dense detectors perform better in spatial equilibrium.
In general, our SP encourages the detector to perform uniformly and well in all zones.

\myPara{Spatial bias in datasets.}
Table \ref{tab:datasets} reports the quantitative detection results on PASCAL VOC and 3 application datasets.
One can see that the detection performance varies across the zones.
The zone nearest to the image border, i.e., $z^{0,1}$, has the consistent lowest detection performance than the others.
In contrast, the central zone $z^{4,5}$ has the highest performance in almost all of these cases.
But one should note that there is an exception such as the helmet dataset, where the ZP in the central zone $z^{4,5}$ is not the best, but the ZP in the zone $z^{3,4}$ is.
This is because the object distribution of the helmet dataset does not satisfy the common centralized photographer's bias.
The helmets and the human heads are more likely to be in the top region of the image (kindly refer to the Appendix).
As a result, the detector also shows a skew zone performance as the object distribution is skew.
The amplitude of spatial disequilibrium also varies for these datasets.
For examples, the variance of ZP is 49.4 on PASCAL VOC but only 3.2 on the helmet dataset.

\subsection{Analysis on SELA}

\myPara{Hyperparameter $\gamma$.}
In Eq. \ref{eq:SELA}, $\gamma$ controls the magnitude of the spatial weight.
A larger $\gamma$ increases more positive samples for objects near the image borders.
As shown in Table \ref{tab:gamma}, we observe that our SELA can achieve a consistent spatial equilibrium improvement (lower variance) for all the options of $\gamma$.
A large $\gamma$ will increase much more positive samples for all zones, leading to a performance drop.
Thus, we set $\gamma$ to 0.2 for PASCAL VOC.
One can see that our SELA can significantly improve the detection performance for the outer zones, e.g., ZP$^{0,1}$ and ZP$^{1,2}$.
To pursue spatial equilibrium, it is reasonable that the performance in the central zone $z^{4,5}$ decreases.
This is because the objects near the image borders receive more supervision signals during training, which relieves the network from paying too much attention to the central objects.
Nevertheless, we believe it is acceptable since $z^{4,5}$ occupies only 4\% image areas and we aim to chase a better spatial equilibrium.
In practice, we set $\gamma=0.1$ for all the other datasets, but it should be noted that there might be a better $\gamma$ for different application scenarios.

\begin{table}[!t]
  \centering
  \small
  \setlength{\tabcolsep}{3.8pt}
  \caption{Evaluation of hyper-parameter $\gamma$.
    SP and ZPs are reported.
    ``Var": the variance of the 5 ZPs.
    $\gamma=0$ denotes the baseline GFocal.
    The results are reported on VOC 07+12.
  }
  \begin{tabular}{c|c|ccccc|cc}
    $\gamma$ &ZP$^{0,5}$&ZP$^{0,1}$&ZP$^{1,2}$&ZP$^{2,3}$&ZP$^{3,4}$&ZP$^{4,5}$&Var& SP \\ \hline
	0  &51.9&31.5&37.7&40.1&\textbf{43.4}&\textbf{52.8}&49.4&37.2 \\ \hline
    0.1&52.2&32.4&38.2&40.0&42.9&52.6&44.1&37.6 \\
    0.2&52.5&\textbf{33.9}&38.6&\textbf{41.5}&43.3&52.5&\textbf{37.9}&\textbf{38.6}\\
    0.3&\textbf{52.6}&33.1&\textbf{39.5}&41.4&\textbf{43.4}&52.2&38.4&38.6\\
    0.4&51.6&31.8 & 36.9 & 40.3 & 43.0 & 51.6 & 43.6 & 37.1 \\
  \end{tabular}
  \label{tab:gamma}
\end{table}

\myPara{Spatial weight.}
One may wonder how would the performance go if we directly loose the selection condition for the positive samples without considering their spatial positions.
Here, we conduct an experiment to investigate the effect of the spatial weight.
The quantitative results are reported in Table \ref{tab:spatial-weight}.
If the spatial weight is set to a constant 1, which means that we directly lower the positive IoU threshold,
\begin{equation}
\textrm{IoU}(\bm{B}^a,\bm{B}^{gt})\geqslant t-\gamma,
\end{equation}
more positive samples will be selected without the spatial discrimination.
One can see that although the ZP$^{0,5}$ increases, the variance of the 5 ZPs is large.
This means subtracting a constant from the positive IoU threshold cannot change the sampling frequency much, since more positive samples are generated in the central zone.
In contrast, our SELA can significantly reduce this variance, achieve a much better spatial equilibrium, and produce better SP.

\begin{table}[!tb]
  \centering
  \small
  \setlength{\tabcolsep}{2.4pt}
  \caption{Analysis of spatial weight.
    SP and ZPs are reported.
    ``Var": the variance of the 5 ZPs. $\gamma=0.2$.
    The results are reported on VOC 07+12.
  }
  \begin{tabular}{c|c|ccccc|cc}
	weight&ZP$^{0,5}$&ZP$^{0,1}$&ZP$^{1,2}$&ZP$^{2,3}$&ZP$^{3,4}$&ZP$^{4,5}$&Var & SP\\ \hline
	0 & 51.9 & 31.5 & 37.7 & 40.1 &\textbf{43.4}&\textbf{52.8}& 49.4 & 37.2 \\
    1 & \textbf{52.5} &31.1 & 38.4 & 41.1 & 43.1 & 52.7 & 49.1 & 37.4 \\
    $\alpha(x^{a},y^{a})$ &\textbf{52.5} & \textbf{33.9} & \textbf{38.6} & \textbf{41.5} & 43.3 & 52.5 & \textbf{37.9} & \textbf{38.6} \\
\end{tabular}
\label{tab:spatial-weight}
\end{table}

\begin{figure*}[!t]
    \centering
    \setlength{\tabcolsep}{1pt}
    \setlength{\abovecaptionskip}{3pt}
    \begin{tabular}{cccccccccc}
        \includegraphics[width=0.96\textwidth]{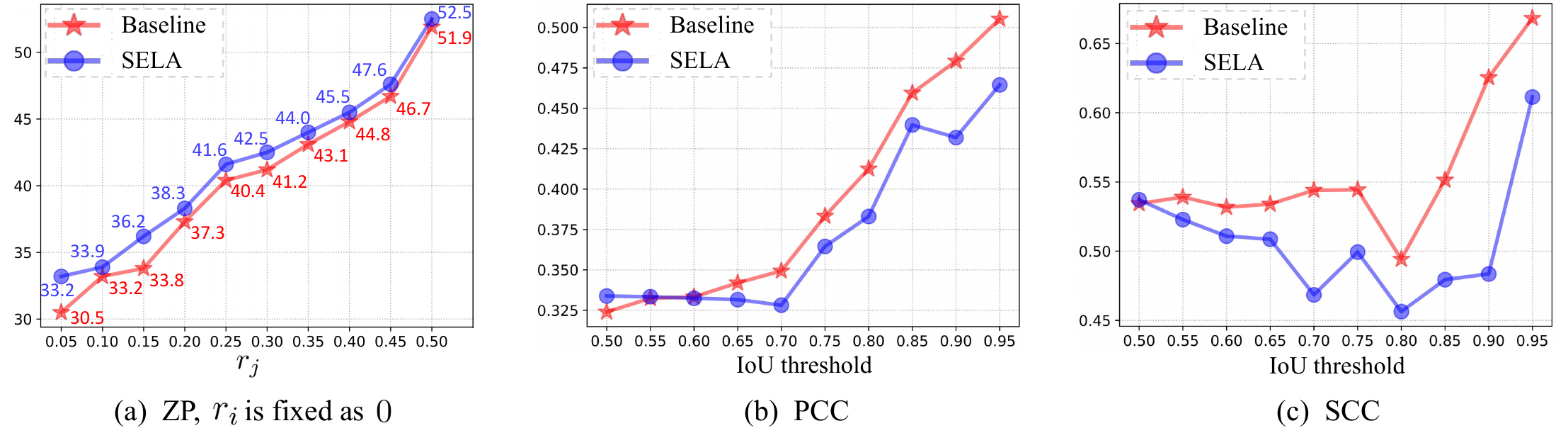}\\
    \end{tabular}
    \vspace{-5pt}
    \caption{(a) ZP against the zone range $(r_i,r_j)$, where $r_i$ is fixed as 0 and $r_j=0.05j$, $j=\{1,2,\cdots,10\}$. (b) Pearson Correlation Coefficient (PCC) between the mZP and the object distribution (center counts) against the IoU threshold. (c) Spearman Correlation Coefficient (SCC) between the mZP and the object distribution against the IoU threshold. Our SELA can substantially reduce these correlations under most of IoU thresholds. The results are reported on VOC 07+12.
    }
    \label{fig:APcurve-PCC-SCC}
\end{figure*}

\myPara{SELA on various datasets.}
Table \ref{tab:datasets} shows us promising results that our SELA can achieve a better spatial equilibrium for object detection.
In particular, we reduce the variance by a large margin in terms of ZP and ZP$_{75}$.
For example, we successfully lower the variance of ZP by -11.5, -8.8, -13.0 and -0.7 on PASCAL VOC, and face mask/fruit/helmet detection.
More importantly, our SELA lifts the SP scores by +1.4, +1.3, +0.8, and +0.8 for the 4 datasets.

\myPara{Performance on border zones.}
As SELA enhances the supervision signal in the border zones, it can be seen from \figref{tab:datasets} that the detection performance can be significantly improved for the border zones $z^{0,1}$.
We also conduct evaluation with a finer zone division, which is the same to the settings of \figref{fig:APcurve}, where $r_i, r_j$ control the range of the annual zones.
As plotted in \figref{fig:APcurve-PCC-SCC}(a), our SELA shows a consistent ZP improvement for all the $r_j$, which is remarkable especially for the border zone, i.e., a small $r_j$.

\myPara{Correlation with object distribution.}
We further provide the correlation between the zone metrics and the object distribution.
We define a finer zone division, which is the same to the division for counting the centers of the objects, i.e., $11\times11$ square zones (see the top of \figref{fig:photographer-bias}).
Then we evaluate the detection performance in the 121 zones one by one.
To quantitatively investigate the correlation between the zone metrics and the object distribution, we calculate the Pearson Correlation Coefficient (PCC) and the Spearman Correlation Coefficient (SCC) between the mZPs and the object distribution of the test set.

As shown in Figs.~\ref{fig:APcurve-PCC-SCC}(b) and (c), we get the following deep reflection about the spatial bias.
We first note that all the PCCs $>0.3$ in \figref{fig:APcurve-PCC-SCC}(b), which indicates that the detection performance is moderately linear correlated with the object distribution.
As a reminder, the PCC only reflects the linear correlation of two given vectors, while it may fail when they are curvilinearly correlated.
In \figref{fig:APcurve-PCC-SCC}(c), the Spearman correlation reflects a higher ranking correlation between the mZPs and the object distribution with all the SCCs $>0.45$.
This illustrates that the detection performance has a similar trend to the object distribution.
In general, our SELA substantially reduces these correlations, indicating a lower correlation with the object distribution.

\begin{table}[!t]
  \centering
  \small
  \setlength{\tabcolsep}{1.7pt}
  \caption{Evaluation of SELA with various backbone networks on
    PASCAL VOC 07+12.
    SP and ZPs are reported.
    ``Var": the variance of the 5 ZPs.
    \textbf{R}: ResNet \cite{ResNet}.
    \textbf{X}: ResNeXt-32x4d-DCN \cite{xie2017aggregated,DCNv2}.
  }
  \begin{tabular}{c|c|c|ccccc|cc}
	Model & SELA & ZP$^{0,5}$ & ZP$^{0,1}$ & ZP$^{1,2}$ & ZP$^{2,3}$ & ZP$^{3,4}$  & ZP$^{4,5}$  & Var & SP  \\ \hline
	\multirow{2}{*}{\textbf{R}-18} & & 51.9 &31.5 & 37.7  & 40.1  & 43.4 & 52.8 & 49.4 & 37.2 \\
    &\checkmark& 52.5 & 33.9 & 38.6 & 41.5 & 43.3 & 52.5 & 37.9 & 38.6\\ \hline
	\multirow{2}{*}{\textbf{R}-50} & & 55.4 & 36.8 & 41.7  & 44.2  & 47.0 & 56.2 & 42.8  & 41.7 \\
    &\checkmark& 55.6 & 39.0 & 42.4 & 45.1 & 46.0 & 54.4 & 22.9 & 42.6\\ \hline
	\multirow{2}{*}{\textbf{X}-101} & &  62.7 & 43.2 & 49.8 & 51.5 & 54.0 & 62.9 & 34.5 & 48.8 \\
    &\checkmark& 63.0 & 44.6 & 50.3 & 51.9 & 53.5 & 62.9 & 27.4 & 49.5\\
  \end{tabular}
  \label{tab:backbone}
\end{table}

\myPara{Generality of SELA.}
We first provide more experiments to verify the effectiveness of SELA on various backbone networks.
Table \ref{tab:backbone} exhibits that our SELA can notably improve the spatial equilibrium for all the 3 backbone networks, i.e., lower variance and better SP.
We also conduct experiments to check out the generality of SELA by incorporating it into 2 more dynamic label assignment algorithms, DW \cite{DW} and DDOD \cite{DDOD2021}.
The principle of applying SELA into these detectors lies in considering the spatial weight when making the criterion rule for determining the positives and negatives.
Here, we adopt a unified implementation, i.e., the cost-sensitive learning approach.
We take the spatial weight term $1+\gamma\alpha(x^{a},y^{a})$ as an additional weight factor for the classification and the bounding box regression losses.
Table \ref{tab:other-detectors} reports the quantitative results of SELA for these 3 dynamic label assignments.
As shown, our method can substantially reduce the ZP variance and improve the SP for the 3 detectors, indicating that a better spatial equilibrium is achieved.
This shows the generalized ability of our method to improve the spatial robustness of detectors without bells and whistles.

\begin{table}[!t]
  \centering
  \small
  \setlength{\tabcolsep}{4pt}
  \caption{Evaluation of SELA (cost-sensitive learning approach)
    on 3 recently popular detectors with dynamic label assignments.
    SP and ZP$^{0,5}$ are reported.
    ``Var": the variance of ZPs over the 5 zones.
  }
  \begin{spacing}{0.86}
  \begin{tabular}{c|c|ccc|ccc}
	\multirow{2}{*}{Method} & \multirow{2}{*}{SELA} &
    \multicolumn{3}{c|}{VOC 07+12} & \multicolumn{3}{c}{COCO 2017 val}\\
	& & ZP$^{0,5}$ & Var & SP & ZP$^{0,5}$ & Var & SP \\ \hline
	\multirow{2}{*}{GFocal \cite{gfocal}} &
    & 51.9 & 49.4 & 37.2 & 40.1 & 14.4 & 35.7 \\
    & \checkmark & 52.1 & 42.4 & 38.3 & 40.0 & 12.4 & 36.0 \\ \hline
	\multirow{2}{*}{DW \cite{DW}} &
    & 52.6 & 42.8 & 38.2 & 42.0 & 18.8 & 37.2 \\
    & \checkmark & 52.6 & 29.3 & 38.9 & 42.2 & 17.0 & 37.5 \\ \hline
	\multirow{2}{*}{DDOD \cite{DDOD2021}} &
    & 50.8 & 25.1 & 37.4 & 41.7 & 16.3 & 36.8 \\
    & \checkmark & 51.3 & 24.1 & 38.1 & 41.6 & 14.4 & 37.1 \\
  \end{tabular}
  \end{spacing}
  \label{tab:other-detectors}
\end{table}

\section{Conclusions and Discussions}\label{sec:conclusion}
In this paper, we study the spatial disequilibrium problem in object detection and show the existence of the spatial bias in modern object detectors.
We present a new zone evaluation protocol to quantitatively evaluate the detection performance in a series of zones.
The outcome is a spatial equilibrium precision (SP), which can comprehensively evaluate the object detectors.
We also present a spatial equilibrium label assignment method to alleviate the spatial disequilibrium problem of modern detectors.
We hope this work could inspire the community to rethink the evaluation of object detectors and stimulate further explorations on the spatial bias.

\myPara{Discussions.}
We notice that the zone division is flexible and we provide a preliminary study on the centralized zone division.
The design of zone division warrants future research.

\myPara{Acknowledgement.}
We would like to thank Professor Ping Wang for her valuable suggestions to this paper.

{
\small
\bibliographystyle{ieee_fullname}
\bibliography{egbib}
}

\newpage

\appendix

\renewcommand{\thesection}{A\arabic{section}}
\renewcommand{\thetable}{A\arabic{table}}
\renewcommand{\thefigure}{A\arabic{figure}}

\setcounter{table}{0}

\begin{figure*}[h]
	\centering
	\setlength{\tabcolsep}{1pt}
	\setlength{\abovecaptionskip}{3pt}
	\begin{tabular}{cccccccccc}
		\includegraphics[width=1\textwidth]{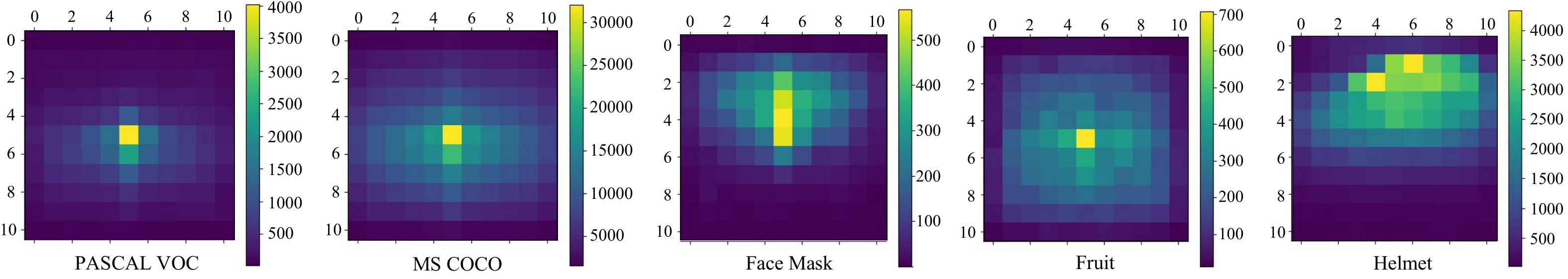}\\
	\end{tabular}
	\caption{The photographer's bias in the 5 object detection datasets. We count the center points for all the ground-truth boxes. The images are divided into $11\times11$ zones.
		}
	\label{fig:5dataset}
\end{figure*}

\begin{figure*}[h]
	\centering
	\setlength{\tabcolsep}{1pt}
	\setlength{\abovecaptionskip}{3pt}
	\begin{tabular}{cccccccccc}
		\includegraphics[width=0.86\textwidth]{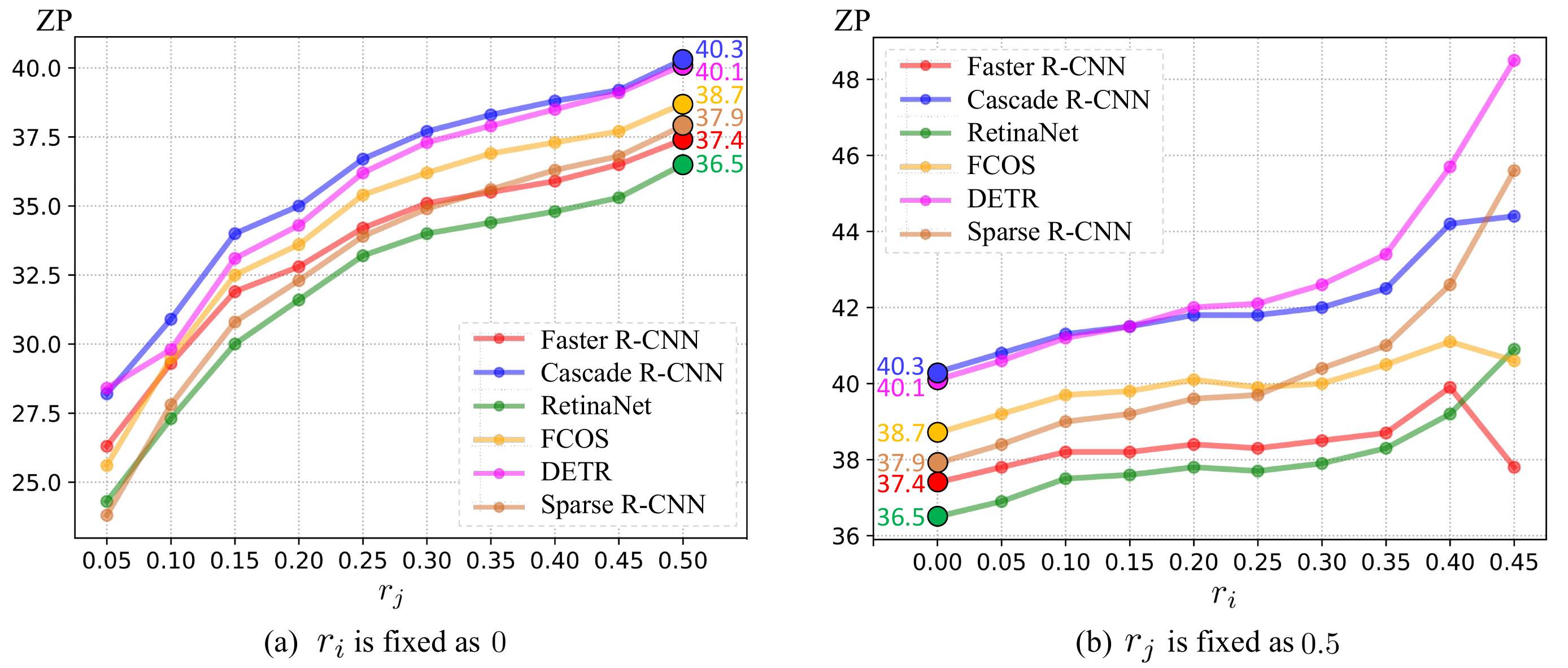}\\
	\end{tabular}
	\caption{The ZP against the zone range $(r_i,r_j)$, where $r_i=0.05i$, $r_j=0.05j$, and $i,j\in\{0,1,2,\cdots,10\}$. All the 6 detectors adopt the popular ResNet-50. The results are reported on the COCO val 2017.
		}
	\label{fig:AP-curve}
\end{figure*}

\section{Datasets}

The experiments are conducted on the following 5 datasets, which are publicly available.
The object distributions can be seen in Fig. \ref{fig:5dataset}.

\myPara{PASCAL VOC \cite{voc}}
is one of the most widely used object detection benchmark under natural scenes, which contains 20 classes.
We adopt the classical 07+12 training and testing protocol, i.e., the train set contains the union of VOC 2007 trainval and VOC 2012 trainval (totally 16551 images) and the test set contains VOC 2007 test (4952 images).

\myPara{MS COCO \cite{coco}}
is another recently popular benchmark with much larger scale, containing 80 classes under natural scenes.
We adopt COCO 2017 train (118K images) for training and COCO 2017 val (5K images) for evaluation.

\myPara{Face mask detection\footnote{\url{https://www.kaggle.com/datasets/parot99/face-mask-detection-yolo-darknet-format}}.}
With COVID-19 raging around the world, face mask detection is a widespread and necessary visual application.
The dataset consists of 5,865 images for training and 1,035 images for testing.
There are 2 classes. One is no mask and the other is mask.

\myPara{Fruit detection\footnote{\url{https://www.kaggle.com/datasets/eunpyohong/fruit-object-detection}}}
is widely used in industrial assembly line sorting and commodity classification.
The dataset consists of 3,836 train images and 639 test images.
11 common fruits are included, e.g., apple, grape, and lemon.

\myPara{Helmet detection\footnote{\url{https://www.kaggle.com/datasets/vodan37/yolo-helmethead/metadata}}}
is a safety vision application that is often used in construction sites to detect whether personnel wear helmets.
It contains 15,887 images for training and 6,902 images for testing.
Two classes, head and helmet, are used.

\section{More Experiments}

\subsection{Zone Evaluation on Detectors.}
We provide zone evaluation to 6 object detectors, including Faster R-CNN \cite{fasterrcnn} (two-stage), Cascade R-CNN \cite{cascadercnn} (multi-stage), DETR \cite{DETR} (transformer-based sparse detector), Sparse R-CNN \cite{sparsercnn} (two-stage sparse detector), RetinaNet \cite{lin2017focal} (one-stage), and FCOS \cite{FCOS} (one-stage).
These detectors adopt the same backbone network ResNet-50 \cite{ResNet}.
For CNN-based detectors, FPN \cite{FPN} is equipped as the neck network and a 1$\times$ training schedule is adopted (12 epochs).
While for DETR, the default 150 epochs training schedule is adopted.
The number of evaluation zones $n$ is set to 10, where the zone ranges are $(r_i,r_j)=(0,0.05j)$ in Fig. \ref{fig:AP-curve}(a), and $(r_i,r_j)=(0.05i,0.5)$ in Fig. \ref{fig:AP-curve} (b), $i\in\{0,1,\cdots,9\}$, $j\in\{1,2,\cdots,10\}$.
Fig. \ref{fig:AP-curve}(a) shows that when $r_j<0.5$, all the ZPs are less than the traditional one ($r_j=0.5$).
On the other hand, Fig. \ref{fig:AP-curve}(b) shows a ZP increasing tendency as the evaluation zones shrink to the image center.
This indicates that the detectors perform better in the central zone than in most border area.
Particularly, we note that the sparse detectors, DETR and Sparse R-CNN favor more in the central zone ($r_i>0.3$), as shown in Fig. \ref{fig:AP-curve}(b).

\begin{table*}[t]
\centering
\tablestyle{4pt}{1.2}
\begin{tabular}{c|c  c  c  | c  c  c |c| c  c  c  | c  c  c   }
				\hline

				\hline
				\multirow{2}{*}{H Flip} & \multicolumn{3}{c|}{Left-0 detector} & \multicolumn{3}{c|}{Left-1 detector} && \multicolumn{3}{c|}{Right-0 detector} & \multicolumn{3}{c}{Right-1 detector}\\
				\cline{2-7}\cline{9-14}
				 & ZP@left  & ZP@right   & ZP@full  & ZP@left & ZP@right  & ZP@full && ZP@left  & ZP@right   & ZP@full  & ZP@left & ZP@right  & ZP@full \\
				\cline{1-7}\cline{9-14}
				& 28.9  &  42.0 & 40.9 & 42.8 & 44.5 &  49.3 && 42.8  &  27.2 & 40.5 & 45.7  &  41.0 & 49.2 \\
				\checkmark & 42.3 & 28.8 & 40.8 & 45.0 &  41.7  & 49.5 && 27.6 &  42.3 & 40.6 & 42.0  &  45.3 & 49.0 \\
				\hline

				\hline
\end{tabular}
\caption{\textbf{Detector's spatial bias}. ``Left-0": discard all the left zone objects. ``Left-1": keep only 1 positive sample for each left zone object. The left/right/full zone ZP is reported. ``Right-0": discard all the right zone objects. ``Right-1": keep only 1 positive sample for each right zone object. The left/right/full zone ZP is reported. ``H Flip": testing with horizontal flip. The detector performs extremely imbalanced between the two zones. Also, increasing training samples for the disfavor zones will shrink the performance gap.}
\label{tab2}
\end{table*}

\subsection{A Close Look at Detectors' Spatial Bias}

As mentioned in Section 1 of the main paper, we found some clues about the detectors' spatial bias when computing the AP scores for different zones.
This object distribution bias makes it difficult for detectors to have consistent performance
across different image zones.
To further demonstrate this,
we create a simple yet heuristic experiment by manually reducing the object supervision signals in a certain zone.
We conduct experiments using the popular GFocal \cite{gfocal} detector under the classic PASCAL VOC 07+12~\cite{voc} protocols.

We first evenly divide the whole image zone into two (left and right) halves.
Then, there are four pipeline settings for comparison:

\noindent
1. ``left-0" detector: we train the network by discarding all the objects whose centers lie in the left zone of the image.

\noindent
2. ``right-0" detector: analogous to ``left-0" detector by discarding the right zone objects.

\noindent
3. ``left-1" detector: we only assign 1 positive sample for every left zone object.

\noindent
4. ``right-1" detector: the opposite settings to ``left-1".

All the four detectors differ only in sampling process.
During training, horizontal flip with a probability of 0.5 is used to ensure that both left and right objects participate in the model training.
The evaluation is conducted on the left zone, the right zone and the whole image separately.

\begin{figure*}[!t]
	\centering
	\setlength{\tabcolsep}{1pt}
	\setlength{\abovecaptionskip}{3pt}
	\begin{tabular}{cccccccccc}
		\includegraphics[width=1\textwidth]{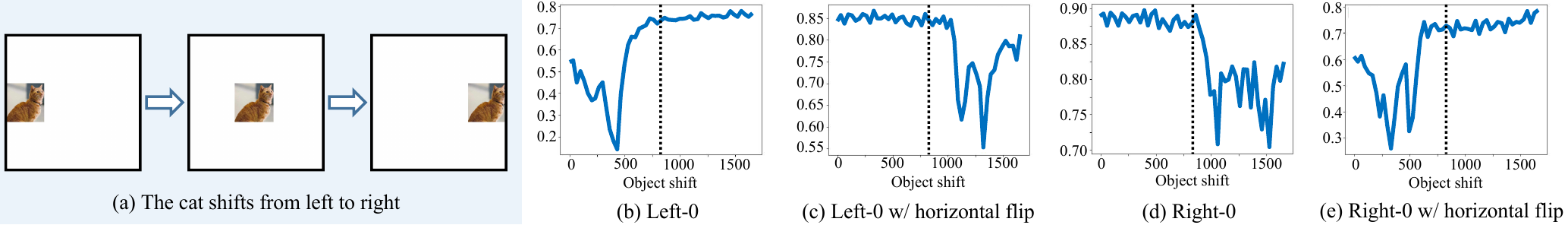}\\
	\end{tabular}
	\caption{\textbf{The cat is not that cat.} The curves (b-e): Product of classification score and IoU (between detected box and the ground-truth box) as the cat shifting from left to right. Imbalance sampling causes severe spatial bias, that the cat in the disfavor zone cannot be detected as good as in the favor zone. If we flip the cat to the favor zone, the detection quality backs to normal immediately.
		}
	\label{fig:cat1}
\end{figure*}

\begin{figure}[!t]
	\centering
	\setlength{\tabcolsep}{1pt}
	\setlength{\abovecaptionskip}{3pt}
	\begin{tabular}{cccccccccc}
		\includegraphics[width=0.46\textwidth]{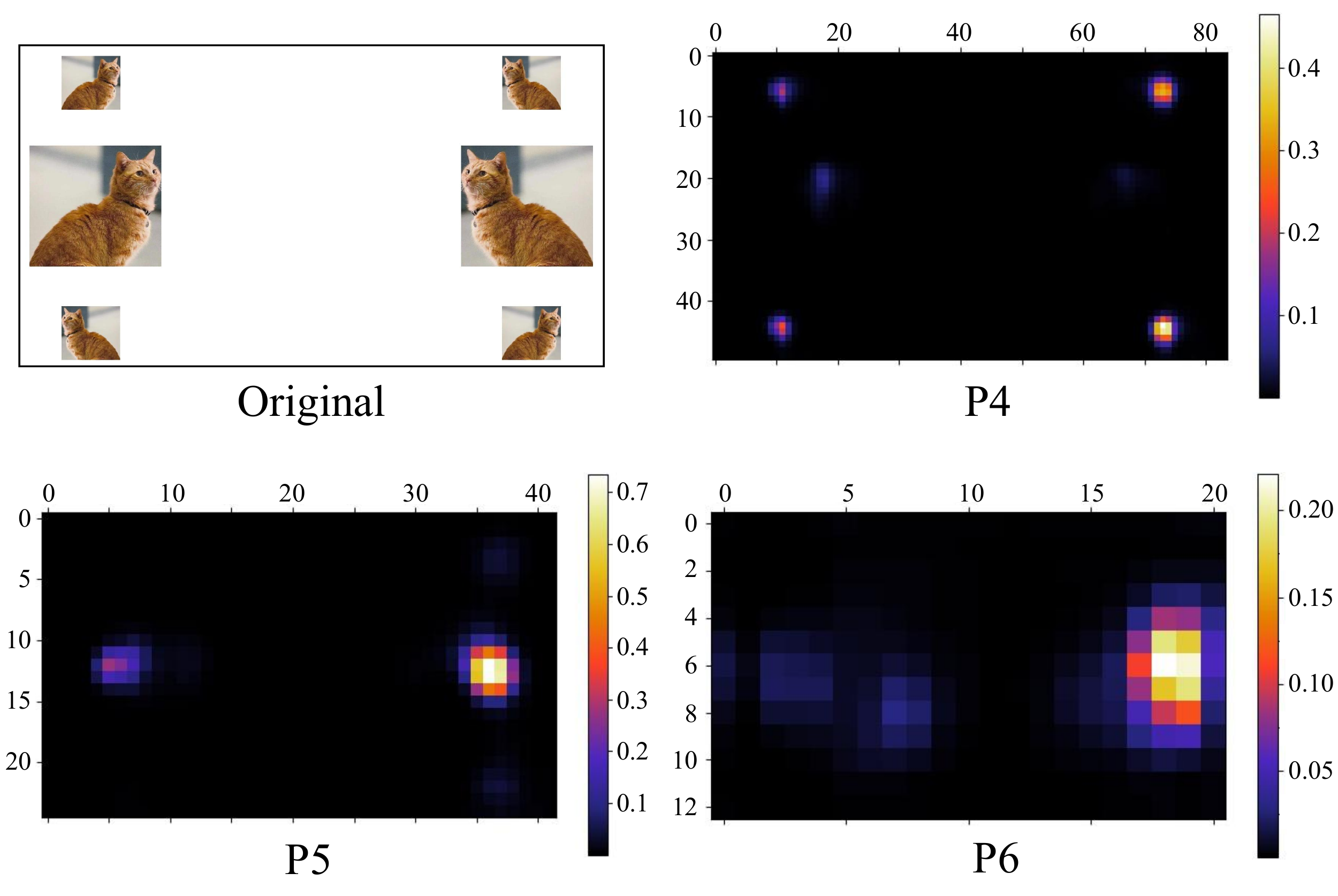}\\
	\end{tabular}
	\caption{Visualization of the class score on P4, P5 and P6 FPN levels. The detector is trained by discarding all the objects whose centers lie in the left zone of the image. The detector produces very weak classification responses for the cats on the left zone.
		}
	\label{fig:cat2}
\end{figure}

Table \ref{tab2} shows the quantitative results and we have the following observations.
\begin{itemize}[noitemsep]
    \item \myPara{Imbalanced sampling of training samples causes severe spatial disequilibrium.}
    It can be seen that the detection performance of the ``left-0" detector in the left zone is very poor, 13.1 ZP worse than that in the right zone.
    This turns out surprisingly that the detector cannot uniformly perform across the zones.
    If we adopt the horizontal flip during testing, it will be completely reversed for the left zone and the right one.
    The same observation can be seen from the ``right-0" detector.
    This implies that the detection performance heavily depends on the positions of objects.
    And the detector is good at detecting objects in the favor zone, simply because it receives much more supervision signals and therefore be endowed much better detection ability during training.
    We also visualize the detection quality in Fig. \ref{fig:cat1}, where the cat shifts from left to right.
    It can be seen that the detection quality will significantly drop when the cat is not at the favor zone.
    If we flip the cat to the favor zone, the detection quality backs to normal immediately.
    In Fig. \ref{fig:cat2}, one can see that the detector produces very weak classification responses for the cats in the disfavor zone.
    Such spatial bias has a great impact on the robustness of detection applications.
    \item \myPara{Traditional metrics fail to capture spatial bias.}
    One can see that the ``left-0" detector produces a 40.9 full map ZP (traditional AP), which is unable to provide a reference for where and how much the performance drops .
    Our zone metrics provide more meaningful information about the detection performance.
    \item \myPara{Increasing training samples for the disfavor zone shrinks the performance gap between zones.}
    Table \ref{tab2} also shows us promising results that the performance gap between the two zones can be significantly shrunk by simply increasing positive samples for the disfavor zones.
    And it should be noted that the performance gap still exists as the sampling remains imbalanced.
\end{itemize}
According to the above analysis, we indicate that the spatial bias not only exists, but also plays a nonnegligible role in gauging the performance of object detectors.

\begin{figure*}[h]
	\centering
	\setlength{\tabcolsep}{1pt}
	\setlength{\abovecaptionskip}{3pt}
	\begin{tabular}{cccccccccc}
		\includegraphics[width=1\textwidth]{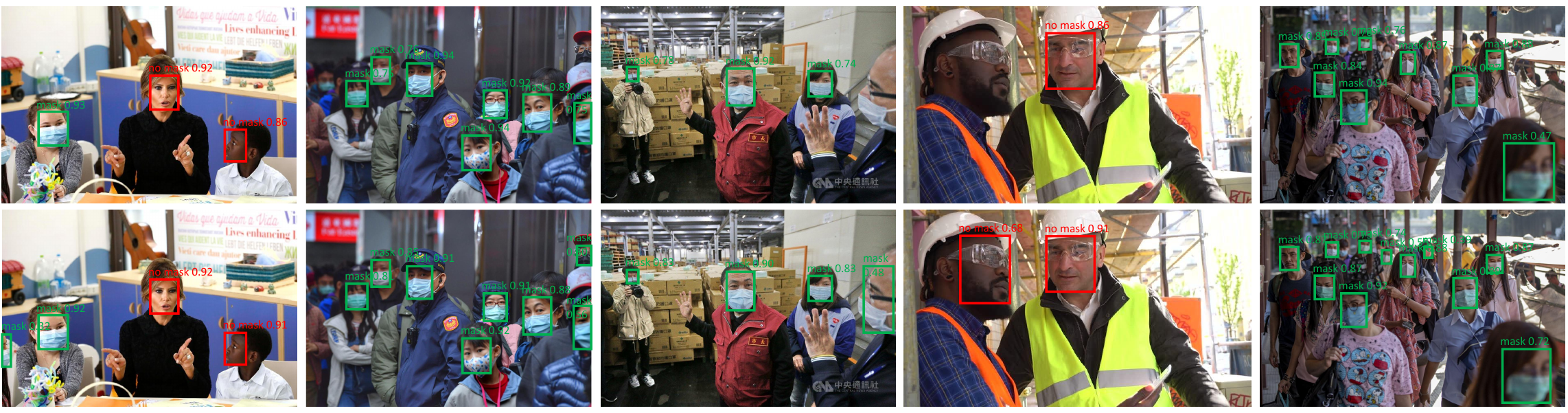}\\
	\end{tabular}
	\caption{Illustration of detection results for GFocal (first row) and GFocal + SELA (second row). Zoom in for a better view.
		}
	\label{fig:base-SELA}
\end{figure*}
\subsection{Visualization of Detection.}
We visualize the detection results of SELA in Fig. \ref{fig:base-SELA}.
Our method can significantly improve the detection performance for the border zone.
We believe the further exploration about spatial equilibrium is clearly worthy and important for the robust detection applications in the future.

\end{document}